\documentclass{article}
\usepackage{amsmath}
\usepackage[
backend=biber,
style=nature,
sorting=none
]{biblatex}
\usepackage{graphicx} 
\usepackage{tabularx}
\usepackage[table]{xcolor}
\usepackage{float}
\usepackage{algpseudocode}
\usepackage{algorithm}
\usepackage[margin=1.1in]{geometry}
\usepackage{authblk}

\title{Predicting Battery Capacity Fade Using Probabilistic Machine Learning Models With and Without Pre-Trained Priors}

\author[1]{Michael J. Kenney}
\author[1]{Katerina G. Malollari}
\author[2, 3]{Sergei V. Kalinin}
\author[3]{Maxim Ziatdinov}

\affil[1]{Amazon Lab126, CA 94089, USA}
\affil[2]{University of Tennessee Knoxville, TN 37909, USA}
\affil[3]{Pacific Northwest National Laboratory, WA 99354, USA}

\begin{document}

\maketitle

\begin{abstract}
Lithium-ion batteries are a key energy storage technology driving revolutions in mobile electronics, electric vehicles and renewable energy storage. Their high energy density, high power output and continually falling cost make them the ideal choice for any application where energy needs to be stored efficiently and reliably. While they have many advantages, a primary concern with lithium-ion batteries is their performance deterioration over time. Capacity retention, a vital performance measure, is frequently utilized to assess whether these batteries have approached their end-of-life. Machine learning (ML) offers a powerful tool for predicting capacity degradation based on past data, and, potentially, prior physical knowledge in the form of simulations or phenomenological models but many previous efforts in this field lack practical approaches for the robust quantification of prediction uncertainty. The degree to which a model's prediction can be trusted is of significant practical importance in situations where consequential decisions need to be made based on battery health decisions. This study explores the efficacy of fully Bayesian machine learning in forecasting battery health with the quantification of uncertainty in its predictions. Specifically, we implemented three probabilistic ML approaches and evaluated the accuracy of their predictions and uncertainty estimates: a standard Gaussian process (GP),  a structured Gaussian process (sGP), and a fully Bayesian neural network (BNN). In typical applications of GP and sGP, their hyperparameters are learned from a single sample while, in contrast, BNNs are typically pre-trained on an existing dataset to learn the weight distributions before being used for inference. This difference in methodology gives the BNN an advantage in learning global trends in a dataset and makes BNNs a good choice when training data is available. However, we show that pre-training can also be leveraged for GP and sGP approaches to learn the prior distributions of the hyperparameters and that in the case of the pre-trained sGP, similar accuracy and improved uncertainty estimation compared to the BNN can be achieved. This approach offers a framework for a broad range of probabilistic machine learning scenarios where past data is available and can be used to learn priors for (hyper)parameters of probabilistic ML models.
\end{abstract}

\section{Introduction}
Accurate prediction of battery health and remaining useful life is important for various applications, ranging from electric vehicles, consumer electronics, renewable energy storage systems and second-life battery pack applications \cite{Hu2020, Harper2019}. As batteries degrade over time due to complex electrochemical processes, their performance deteriorates, leading to reduced capacity and efficiency \cite{Attia_2022}. Early detection of battery degradation can enable timely maintenance, replacement, or optimization strategies, thereby enhancing system reliability and cost-effectiveness. Traditional battery modeling approaches often rely on physics-based models or empirical models derived from experimental data \cite{Ramadesigan_2012, WAAG2014321}. However, the physics underlying battery aging depends on multiple interrelated degradation mechanisms and current battery models are unable to accurately capture these systems. Empirical models benefit from operating at a higher level of abstraction but require assumptions, typically based on domain knowledge and experience, that make the quantification of uncertainty difficult. Alternatively, data-driven approaches, such as machine learning (ML) techniques, offer promising solutions for capturing the complex nonlinear relationships between battery operating conditions, usage patterns, and degradation mechanisms \cite{Aykol_2021}. However, many current ML approaches such as neural networks (NNs) are not well-suited to produce reliable uncertainty estimates \cite{Thelen2024, ovadia2019can}. This weakness limits the utility of ML models to make predictions that can be evaluated objectively in order to take concrete actions in use cases where poor predictions can have significant safety or financial repercussions.

In this study, we examine the impact of pre-training on accuracy and reliability of three probabilistic ML approaches — Gaussian Processes (GP), Structured Gaussian Processes (sGP), and Bayesian Neural Networks (BNN) — for predicting battery health. Traditionally, GP models \cite{rasmussenbook} are trained from scratch with non/weakly-informative priors for each individual performance curve and are utilized to forecast performance trends based solely on the data from that curve, without leveraging broader data trends. While this approach is effective in scenarios lacking historical data or simulations, it may not fully utilize already available data that could reveal overarching trends. The pre-training stage modifies this by using a separate preliminary dataset to establish the prior distributions for the GP and sGP models. Similarly, Bayesian Neural Networks (BNNs) can be employed either in a single-shot approach as direct substitutes for standard GPs or in a pre-training approach, where posterior distributions over NN weights learned from prior data are used to generate predictive distributions for new inputs. This work explores whether pre-training these probabilistic models enables better generalization and more accurate uncertainty quantification for battery performance predictions.

\begin{figure}[ht]
    \centering
    \includegraphics[width=1\linewidth]{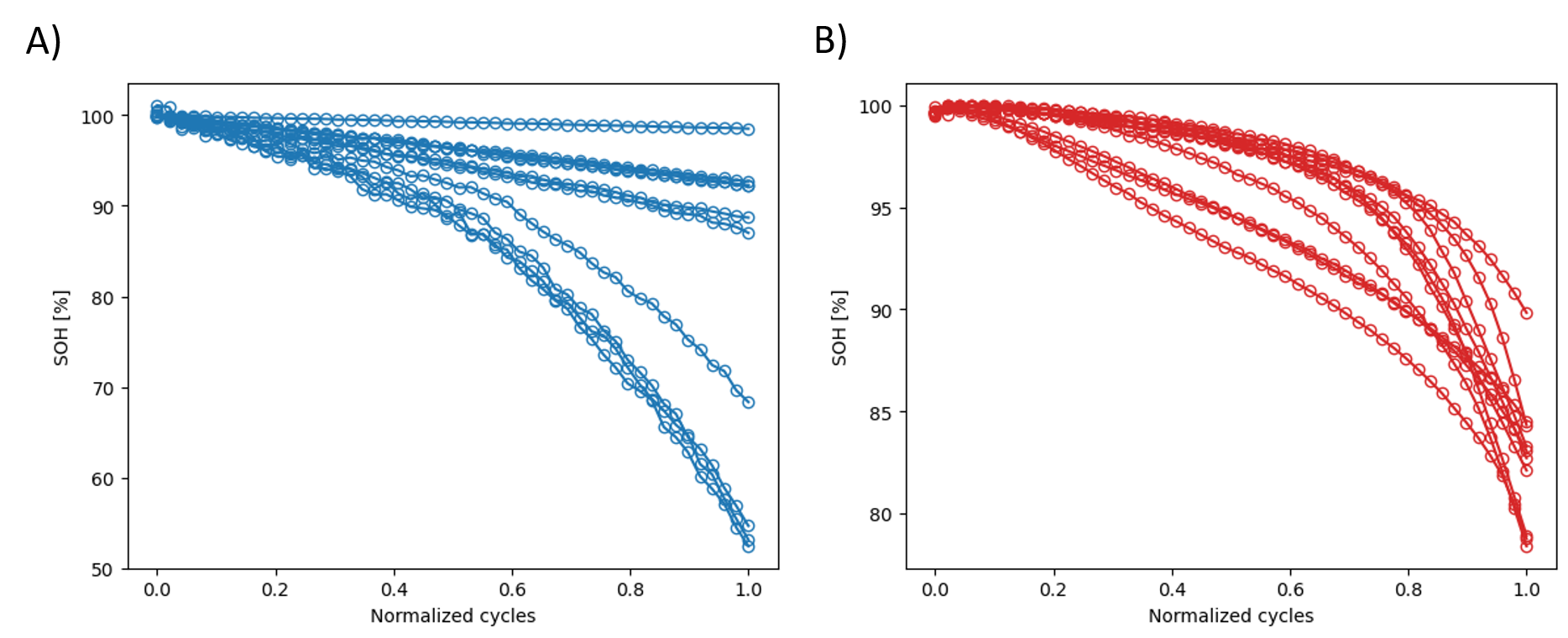}
    \caption{(A) 10 randomly selected curves (out of 250) from the simulated dataset and (B) 10 randomly selected curves (out of 125) from the experimental dataset.}
    \label{fig:sim-exp-data-comparison}
\end{figure}

\section{Methods} 
\subsection{Gaussian Process}
A standard GP can be described as a probabilistic model of the form

\begin{equation}
    \begin{aligned}
        \text{Latent Function:} & \quad f(x) \sim \mathcal{GP}(m(x; \phi), k(x, x'; \theta)) \\
        \text{Observational Noise:} & \quad \epsilon \sim \mathcal{N}(0, \sigma^2)\\
        \text{Observations:} & \quad y = f(x) + \epsilon
    \end{aligned}
\end{equation}
where \( f(x) \) is the latent function, \( m(x) \) is the mean function parameterized by \( \phi \), \( k(x, x'; \theta) \) is the covariance function (or kernel) with hyperparameters \( \theta \), and \( \epsilon \) is zero-centered Gaussian noise with variance \( \sigma^2 \).

We train the GP by finding the hyperparameters \( \theta \) and noise variance \( \sigma^2 \) and, if applicable, parameters \( \phi \) that maximize the marginal likelihood of the observed data \( \{X, y\} \). After training, we compute the predictive mean \( \mu_* \) and variance \( \Sigma_* \) for new input points \( X_* \) as

\begin{equation}
    \mu_* = m_\phi(X_*) + K_\theta(X_*, X) K_\theta^{-1} (y - m_\phi(X))
    \label{eq:predictive_mean}
\end{equation}
\begin{equation}
    \Sigma_* = K_\theta(X_*, X_*) - K_\theta(X_*, X) K_\theta^{-1} K_\theta(X, X_*)
    \label{eq:predictive_var}
\end{equation}
where \( K_\theta := K_\theta(X, X) + \sigma^2 I \).

In a fully Bayesian GP, we also place priors over the kernel hyperparameters \( \theta \) and the noise variance \( \sigma^2 \), as well over the parameters \( \phi \) of any non-zero mean function, thus incorporating uncertainty in these parameters into the GP model:

\begin{equation}
    \begin{aligned}
        \text{Kernel Hyperparameters:} & \quad \theta \sim p(\theta) \\
        \text{Mean Function Parameters:} & \quad \phi \sim p(\phi) \\
        \text{Noise Variance:} & \quad \sigma^2 \sim p(\sigma^2) \\
    \end{aligned}
\end{equation}

We then use advanced Markov Chain Monte Carlo sampling techniques, such as Hamiltonian Monte Carlo (HMC) \cite{hmc_intro}, to sample directly from the posterior distributions of the hyperparameters \( \theta \) and \( \sigma^2 \). The predictive distribution at new points is then obtained by averaging over these posterior samples. Specifically, the overall predictive mean at new points is given by

\begin{equation}
    \mu_{\text{post}} = \frac{1}{N} \sum_{i=1}^N \mu_*^{(i)}
\end{equation}
where \( N \) is the total number of the HMC samples and \( \mu_*^{(i)} \) is the predictive mean for the \( i \)-th posterior sample, defined using the standard GP predictive mean formula from Equation \ref{eq:predictive_mean}.

The overall predictive variance at new points \( X_* \) can be obtained by combining the individual predictive variances and the variance across the posterior predictive means:

\begin{equation}
    \Sigma_{\text{post}}  = \frac{1}{N} \sum_{i=1}^N \Sigma_*^{(i)} + \text{Var}\left(\mu_*^{(i)}\right)
\end{equation}
where \( \Sigma_*^{(i)} \) is the predictive variance for the \( i \)-th posterior sample, defined using the standard GP predictive variance formula from Equation \ref{eq:predictive_var}. A standard Matern kernel \cite{rasmussenbook} was used for all GPs in this study.

\subsection{Structured Gaussian Process}
Traditionally, the prior mean function \( m \) in GP is set to zero. However, it has been shown that replacing zero mean function with a parametric model of expected system’s behavior can lead to significant improvement in predictive accuracy and the efficacy of GP and GP-based active learning and optimization techniques \cite{ziatdinov_hypothesis_learning, Ziatdinov_ACS_Nano, noack2021gaussian}. This has been demonstrated in the approximated and fully Bayesian modes. In the latter scenario, referred to as structured GP \cite{Ziatdinov_MLST2022}, the posterior distributions over the model parameters result in a more refined GP predictive uncertainties.

In this work, we have chosen the phenomenological model introduced in \cite{Diao2019} as our prior mean function, 
\begin{equation}
m(x) = 1 - A_1x^{b_1} - A_2x^{b_2}
    \label{eq:mean}
\end{equation}
with either half-normal or log-normal priors over its parameters.

\subsection{Bayesian Neural Network}
In BNNs, fixed weights \( w \) of a neural network \( g \) are replaced with prior probability distributions, \(p( w \)), which are updated through Bayes' rule to derive the corresponding posterior distributions after observing data \cite{neal1996bnn, izmailov2021bayesian, ziatdinov2024active}. This approach mitigates common limitations of conventional neural networks, such as overfitting and lack of uncertainty quantification. A homoskedastic BNN can be defined as follows:

\begin{equation}
  \begin{aligned}
    \text{Weights:} & \quad w \sim p(w) \\
    \text{Noise Variance:} & \quad \sigma^2 \sim p(\sigma^2) \\
    \text{Observations:} & \quad y \sim \mathcal{N}(g(x; w), \sigma^2)
  \end{aligned}
\end{equation}

The parameters of this fully stochastic neural network can be inferred using HMC. The predictive mean and uncertainty at new data points are then given by:

\begin{align}
\mu_{post} &= \frac{1}{N} \sum_{i=1}^N g(X_*; w_{\text{post}}^{(i)}) \label{eq:mean_bnn} \\
\Sigma_{post} &= \frac{1}{N} \sum_{i=1}^N (y_*^{(i)} - \mu_{post})^2 \label{eq:uncertainty_bnn}
\end{align}
where

\begin{equation}
y_*^{(i)} \sim \mathcal{N}(g(X_*; w^{(i)}), \sigma^{2(i)})
\end{equation}
are samples from the model posterior at new inputs \( X_* \).

The GP, structured GP, and BNN models were all implemented through GPax open-source software package.\footnote{https://github.com/ziatdinovmax/gpax}

\subsection{Physics-Based Model Simulations}
Simulated data was generated using the PyBaMM \cite{Tranter2022} implementation of a physics-based single particle model with electrolyte (SPMe). Degradation physics of ethylene carbonate reaction limited SEI growth, x-averaged SEI film resistance, partially reversible lithium plating, swelling particle mechanics and stress-driven loss of active material. The PyBaMM parameter set OKane2022 was used as starting point for generating capacity vs cycle data for nickel-cobalt-manganese (NCM) with graphite. Key parameters were randomly varied over pre-defined ranges for each sample to produce a dataset with significant diversity (Table \ref{tab:pbm-params}). 

\begin{table}[h]
    \fontsize{8}{12}\selectfont
    \centering
    \begin{tabular}{ |c|c|c| } 
         \hline
         \rowcolor{lightgray}
         \textbf{Parameter}& \textbf{Min} & \textbf{Max} \\
         \hline
         Ambient T (K) & 280 & 333 \\
         \hline
         SEI kinetic rate constant (m/s) & 1e-18 & 5e-17 \\
         \hline
         EC diffusivity ($m^2$/s) & 1e-21 & 3e-18 \\
         \hline
         Li plating kinetic rate constant (m/s) & 1e-11 & 1e-10 \\
         \hline
         Positive/negative electrode LAM constant proportional term (1/s) & 3e-5 & 5e-5 \\
         \hline
         Initial outer SEI thickness (m) & 1e-9 & 20e-9 \\
         \hline
         Max charge voltage (V) & 3.9 & 4.2 \\
         \hline
         Min discharge voltage (V) & 2.5 & 3.3 \\
         \hline
         Charging C-rate & 0.1 & 0.5 \\
         \hline
         Discharging C-rate & 0.1 & 0.5 \\
         \hline
    \end{tabular}
    \caption{Parameters and ranges used to generate the synthetic battery capacity vs cycle data for NCM-graphite cell described in \cite{OKane_2022}, \cite{Chen_2020}.}
    \label{tab:pbm-params}
\end{table}

\subsection{Pre-training and data processing}
\begin{figure}[ht]
    \centering
    \includegraphics[width=1.0\linewidth]{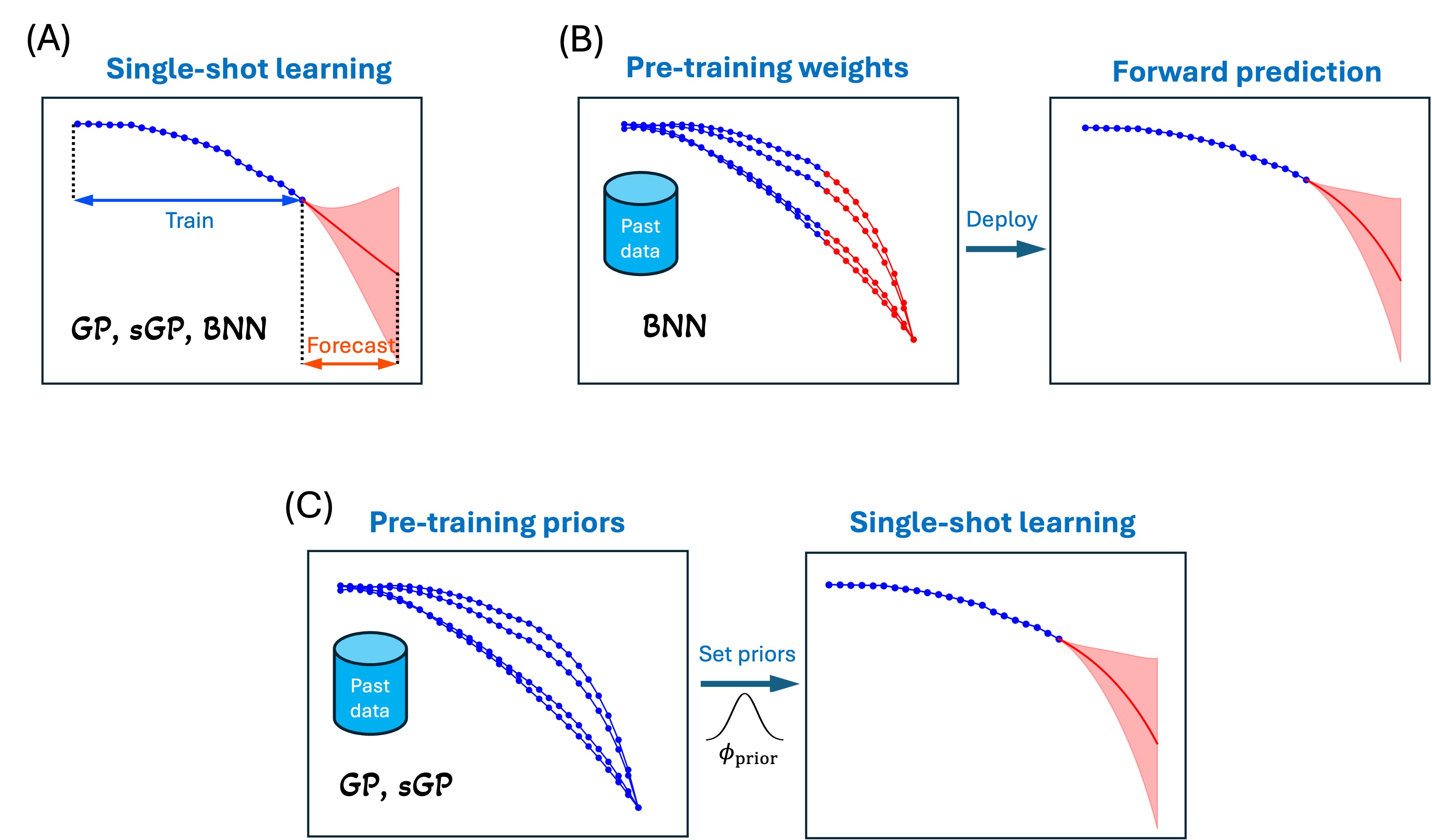}
    \caption{Schematic illustration of three different approaches for forecasting battery performance. (A) Model is trained on the part of the curve for which data is available and is used to forecast battery state of health on the remaining part. (B) Model weights are pre-trained off-line using available data to forecast the battery state of health from partially measured data. It is then deployed for new curves as is. (C) Model priors are pre-trained offline using available data and used to set-up new priors for a single-shot model from A.}
    \label{fig:Fig2}
\end{figure}

Traditionally, GP and sGP are used without pre-training and the hyperparameters are learned from only the observed data on the sample that the prediction will be carried out on \cite{RICHARDSON2017209, THELEN2022119624, RICHARDSON2019320}. This can be described as a 'single-shot' modeling (Figure\ref{fig:Fig2}(A)) and it has the benefit of not requiring a large training dataset, which makes GP-based methods particularly useful when data is limited. This is in contrast to the traditional pre-training approach for neural networks, where a separate dataset is used to train a model prior to deployment on new data (Figure\ref{fig:Fig2}(B)). Here, we explored the potential advantages of pre-training the GP and sGP on the already available data. Towards this goal, we train GP and sGP models on each historical data curve, storing the inferred samples of (s)GP parameters after each training. From these samples, we calculate the overall mean and variance of each parameter, and use those values to set priors when initializing a new, single-shot GP model (Figure\ref{fig:Fig2}(C)). This new model is trained on the measured section of the new performance curve, and it subsequently predicts the unmeasured part, leveraging the insights gained from the historical data to inform these predictions. The detailed description of this method is provided in Algoritm 1. The underlying assumption is that these pre-trained priors can enhance the model's predictive power by providing a more informed starting point, which is particularly valuable when only a limited portion of new data is available for training. 

To test this approach, we utilized two sets of capacity vs cycle data - the fast charging experimental dataset (125 total curves) by Severson \textit{et al.} \cite{Severson2019}  and a simulated dataset generated with PyBaMM (250 total curves), illustrated in Figure \ref{fig:sim-exp-data-comparison}. The capacity curves were normalized to the maximum capacity of each curve (metric commonly used as a battery state-of-health (SOH)) and slightly interpolated or trimmed, such that each curve would consist of 50 points over an x-space from 0 to 1. This processing makes training and testing with the various models more consistent and makes comparison straightforward.

After processing, the curves in both datasets were randomly divided into a train-test split of 80\%-20\%. For GP and sGP pre-training, the models were fit on the entire training dataset without running predictions. At the inference phase, the model is fit to the observed test data, also referred to as the context, and a forecast is made based on this data. Conversely, the BNN models were trained by attempting to forecast the unknown parts of each individual curve from given contexts, adjusting the distributions over weights to minimize the overall error. The learned distributions of weights were then used to obtain a forecast for new data at the inference stage, without any additional fitting.

\begin{algorithm}[H]
\caption{(Structured) Gaussian Process with Pre-Trained Priors}
\begin{algorithmic}[1]

\State \textbf{Input:} Historical data: $\{\mathcal{D}_i\}_{i=1}^N$ (each $\mathcal{D}_i$ is a performance curve)
\State \textbf{Input:} New performance curve: $\mathcal{D}_{\text{new}}$  (consists of $\mathcal{D}_{\text{measured}}$ and $\mathcal{D}_{\text{unmeasured}}$)

\For{each curve $\mathcal{D}_i$ in $\{\mathcal{D}_i\}_{i=1}^N$}
    \State Use MCMC to train (s)GP model $\mathcal{M}_i$ on $\mathcal{D}_i$
    \State Store MCMC samples of (s)GP parameters: $\{\mu_{ij}, \sigma_{ij}^2\}_{j=1}^{J}$ for $\mathcal{M}_i$
\EndFor

\State Compute mean $\mu_{\text{pt}}$ and variance $\sigma^2_{\text{pt}}$ from stored MCMC samples:
\State \quad $\mu_{\text{pt}} = \frac{1}{N} \sum_{i=1}^{N} \left( \frac{1}{J} \sum_{j=1}^{J} \mu_{ij} \right)$
\State \quad $\sigma^2_{\text{pt}} = \frac{1}{N} \sum_{i=1}^{N} \left( \frac{1}{J} \sum_{j=1}^{J} \sigma_{ij}^2 \right)$

\State Initialize new (s)GP model $\mathcal{M}_{\text{new}}$ with priors set to $\mu_{\text{pt}}$ and $\sigma^2_{\text{pt}}$
\State Train $\mathcal{M}_{\text{new}}$ on $\mathcal{D}_{\text{measured}}$
\State Use $\mathcal{M}_{\text{new}}$ to predict $\mathcal{D}_{\text{unmeasured}}$

\end{algorithmic}
\end{algorithm}

\subsection{Model inference and evaluation}
Predictive accuracy and uncertainty quantification were evaluated using mean absolute percentage error (MAPE) and negative log predictive density (NLPD). Only the last five points of each curve were used to calculate these values as the ending SOH behavior is of the most value. 

The MAPE is defined as 

\begin{equation}
    \text{MAPE} = \frac{1}{N} \sum_{i=1}^{N} \left|\frac{y_i - F_i}{y_i}\right| \times 100\%
    \label{eq:mape}
\end{equation}
where N is the number of points, $y_i$ represents the actual values and $F_i$ represents the forecasted values. The NLPD is defined as 

\begin{equation}
    \text{NLPD} = -\frac{1}{N} \sum_{i=1}^N \left[ -\frac{1}{2} \log(2\pi \sigma_i^2) - \frac{(y_i - F_i)^2}{2 \sigma_i^2} \right]
    \label{eq:nlpd}
\end{equation}
where $\sigma_i^2$ is the predicted variance. While MAPE provides a direct measure of error magnitude in predictions, NLPD also assesses how well the predictive uncertainties are estimated \cite{JMLR_gp_nlpd}. It can thus be used to assess  model's ability to generate useful probability distributions of outcomes, which is crucial for decision-making under uncertainty. Smaller values of MAPE and NLPD indicate better model performance.

\section{Results and Discussion}
\subsection{Probabilistic forecasting with and without pre-training}

\begin{figure}[ht]
    \centering
    \includegraphics[width=1\linewidth]{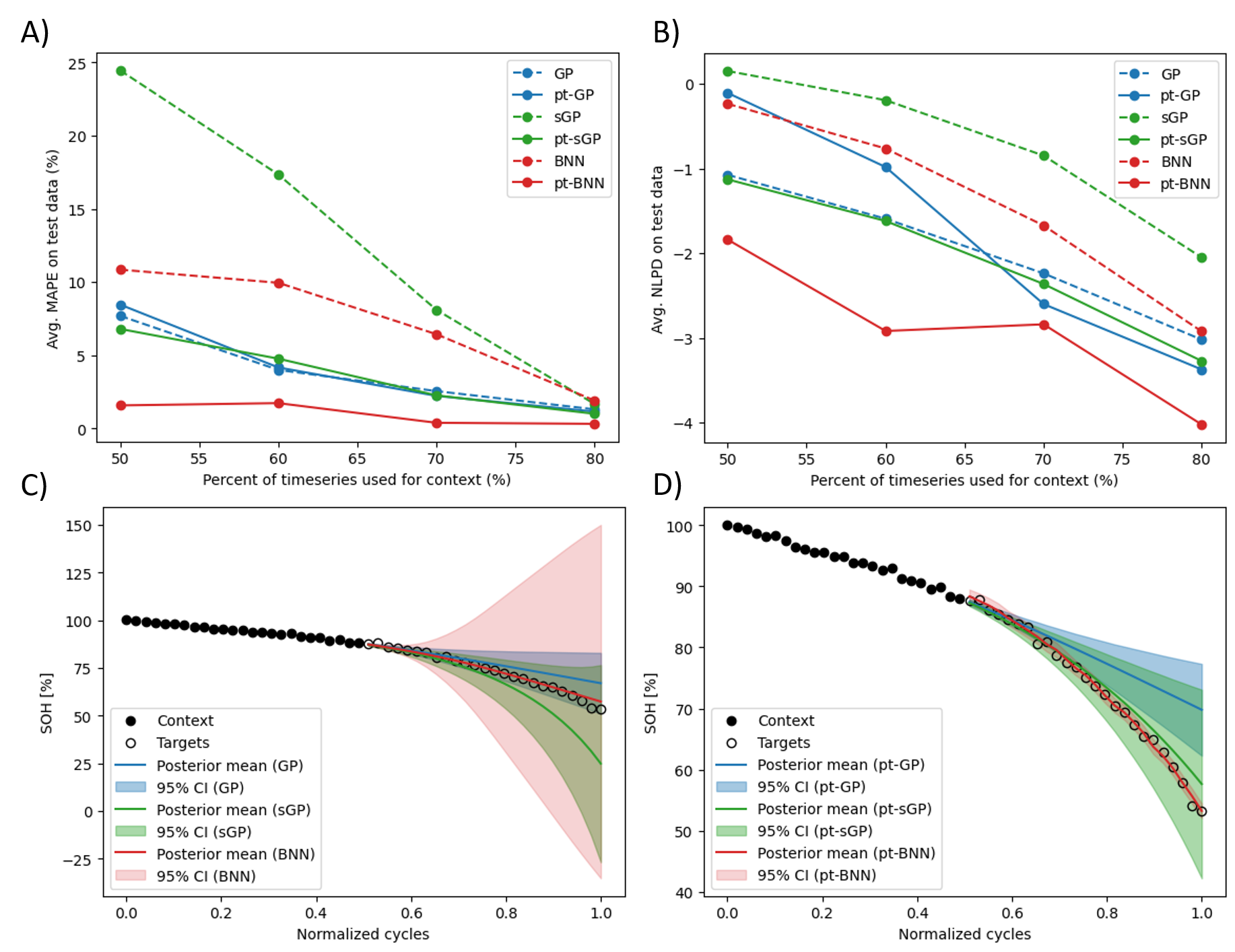}
    \caption{Average MAPE (A) and average NLPD (B) for all six models on the simulated test dataset from 50\%-80\% context length. SOH vs normalized cycles with posterior mean and 95\% confidence intervals on simulated testing data for models without pre-training (C) and with pre-training (D).}
    \label{fig:sim-results}
\end{figure}

\begin{figure}[ht]
    \centering
    \includegraphics[width=1\linewidth]{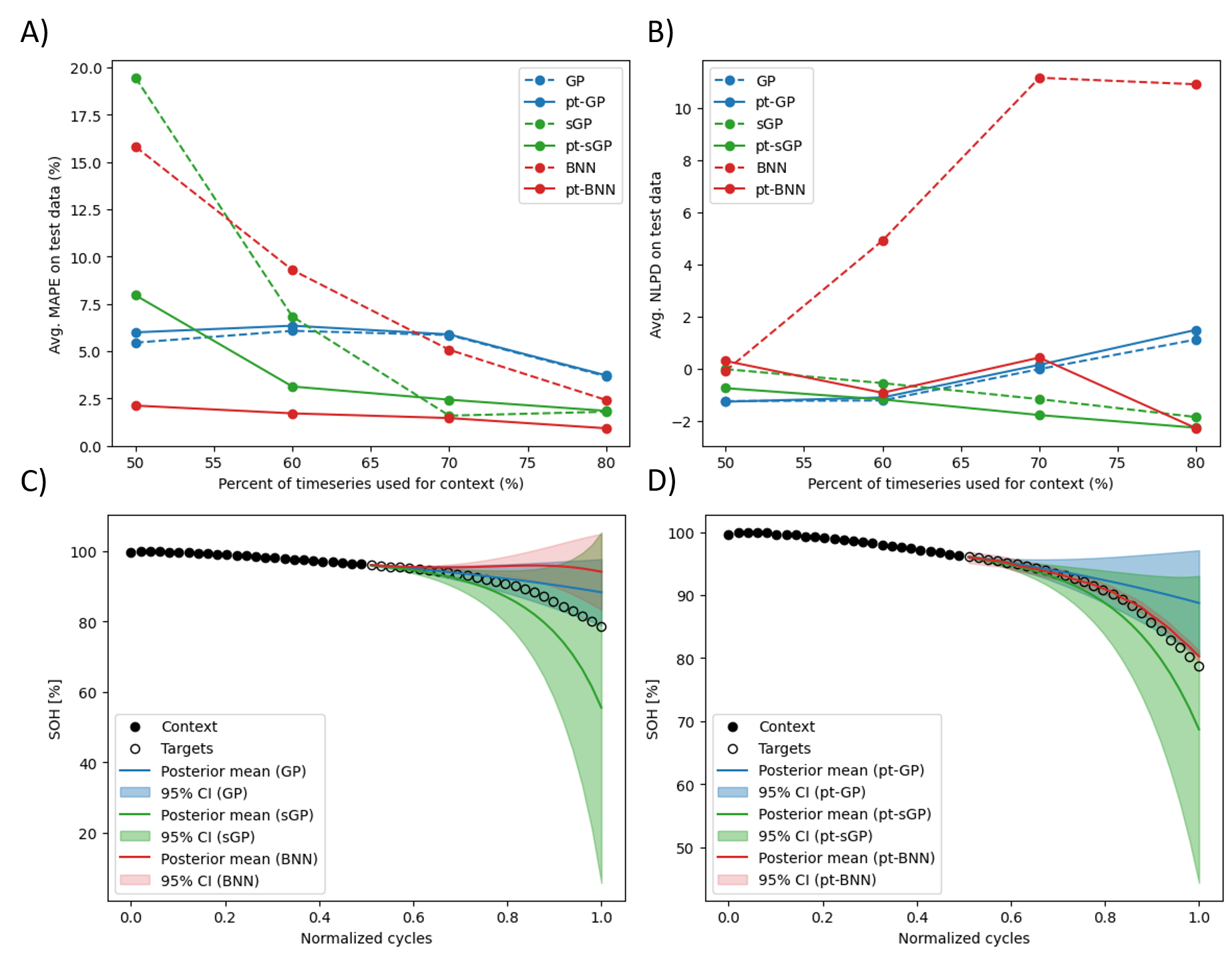}
    \caption{Average MAPE (A) and average NLPD (B) for all six models on the experimental test dataset from 50\%-80\% context length. SOH vs normalized cycles with posterior mean and 95\% confidence intervals on experimental testing data for models without pre-training (C) and with pre-training (D).}
    \label{fig:exp_results}
\end{figure}

The impact of pre-training on the three model types can be graphically seen in figures \ref{fig:sim-results} and \ref{fig:exp_results} with the results tabulated in tables \ref{tab:MAPE_table_sim}, \ref{tab:NLPD_table_sim}, \ref{tab:MAPE_table_exp} and \ref{tab:NLPD_table_exp}. On the simulated dataset without pre-training (Fig. \ref{fig:sim-results}A and C), the GP shows the best performance while the sGP and BNN have large MAPE values until they all converge between 1-2\% MAPE at 80\% context. The BNN and sGP also have wide confidence intervals with the BNN extending the uncertainty to unrealistic values that do not reflect the physical realities of batteries \cite{Attia_2022, BatteryDegradationWYNTK}. After pre-training we can see that the mean predictions are significantly improved for the BNN and sGP with the impact being smaller on the GP. The confidence intervals are significantly narrower and now are consistent with what is reasonably expected by battery physics. 

Note that for sGP, we started with lognormal priors on the $b_1$ and $b_2$ in Equation \ref{eq:mean} to give the GP more flexibility when learning the distributions during pre-training and also to restrict the priors to positive values such that the general curve shape would be maintained (i.e. linear followed by accelerating capacity fade). After pre-training, a normal distribution is used to leverage the learned mean value and standard deviation from the previous step. 

\begin{table}[h]
    \fontsize{8}{12}\selectfont
    \centering
    \begin{tabular}{ |c|c|c|c|c| } 
         \hline
         \rowcolor{lightgray}
         \textbf{Model}& \textbf{50\% context} & \textbf{60\% context} & \textbf{70\% context} & \textbf{80\% context} \\
         \hline
         GP & 7.69\% & 3.99\% & 2.54\% & 1.30\% \\
         \hline
         pt-GP & 8.45\% & 4.16\% & 2.22\% & 1.14\%  \\
         \hline
         sGP & 24.46\% & 17.34\% & 8.09\% & 1.66\% \\
         \hline
         pt-sGP & 6.79\% & 4.76\% & 2.25\% & 0.99\% \\
         \hline
         BNN & 10.84\% & 9.95\% & 6.43\% & 1.89\% \\
         \hline
         pt-BNN & 1.56\% & 1.72\% & 0.38\% & 0.31\% \\
         \hline
    \end{tabular}
    \caption{Average MAPE for all models with and without pre-training on simulated data.}
    \label{tab:MAPE_table_sim}
\end{table}

\begin{table}[h]
    \fontsize{8}{12}\selectfont
    \centering
    \begin{tabular}{ |c|c|c|c|c| } 
         \hline
         \rowcolor{lightgray}
         \textbf{Model}& \textbf{50\% context} & \textbf{60\% context} & \textbf{70\% context} & \textbf{80\% context} \\
         \hline
         GP & -1.07 & -1.59  & -2.23  &  -3.02 \\
         \hline
         pt-GP & -0.11  & -0.98 & -2.60  & -3.37 \\
         \hline
         sGP & 0.15  & -0.19 & -0.85 & -2.05 \\
         \hline
         pt-sGP & -1.13 & -1.62 & -2.36 & -3.27 \\
         \hline
         BNN & -0.24 & -0.76 & -1.67 & -2.92 \\
         \hline
         pt-BNN & -1.84 & -2.92 & -2.84 & -4.02 \\
         \hline
    \end{tabular}
    \caption{Average NLPD for all models with and without pre-training on simulated data.}
    \label{tab:NLPD_table_sim}
\end{table}

Similar behavior can be seen on the experimental dataset for the GP and sGP without pre-training (Fig. \ref{fig:exp_results}C). Interestingly, the BNN's mean prediction is significantly worse compared to the simulated dataset but the confidence intervals are much narrower. This is also clear when examining the tabulated data where we can see that the average MAPE at 50\% context for the BNN (without pre-training) is 10.84\% on the simulated data vs 15.81\% on the experimental data. After pre-training (Fig. \ref{fig:exp_results}D), all three models perform better with the BNN and sGP showing the largest MAPE improvements and the GP showing similar.

\subsection{Encoding information in pre-training}

Pre-training had a significant beneficial impact for both the sGP and the BNN. For example, the MAPE of the sGP at 50\% context improved from 24.46\% without pre-training to 6.79\% with pre-training on the simulated dataset. The experimental results show similar behavior with the sGP MAPE at 50\% context improving from 19.46\% to 7.96\% with pre-training. Similarly, the BNN MAPE at 50\% context on the simulated dataset decreased from 10.84\% to 1.56\% with pre-training and, on the experimental dataset, decreased from 15.81\% to 2.13\% with pre-training. 

The dramatic improvements in MAPE for sGP, compared to the standard GP, can be attributed to their semi-parametric versus non-parametric nature, respectively. Indeed, the standard GP is a non-parametric model where the choice of kernel has a more significant impact on the model's performance than the exact values of its hyperparameters. Practically, as long as the hyperparameters are within a reasonable range, the kernel's overall behavior dictates the model performance more strongly. Furthermore, the marginal likelihood surface for hyperparameters is often quite smooth, meaning that small changes in hyperparameters do not drastically alter the model's predictions, leading to a limited impact of fine-tuning.

Structured Gaussian Process, on the other hand, is a semi-parametric model where a structured parametric model is used as the mean function, capturing known patterns and trends in the data based on prior knowledge. The GP kernel effectively models the residuals, or deviations, from this mean function, allowing for additional flexibility and capturing any patterns not explained by the mean function. This structured approach starts with a better approximation of the underlying function, and if the past data is representative, the priors provide even a better starting point, thereby improving the model's performance. For BNNs, the impact of pre-trained versus single-shot models is even more pronounced due to their fully parametric nature, enabling for the direct encoding of past knowledge into the probabilistic weights of the model.

\subsection{Impact of context window length}

A key consideration for the performance of capacity forecasting models is how much of the data needs to be known in order to make an accurate prediction of the end behavior. To investigate this aspect of the model behavior, we tested the accuracy and uncertainty quantification on 50\%, 60\% ,70\% and 80\% of each sample as the input context. Together with the result tables, figures \ref{fig:sim-results} and \ref{fig:exp_results} illustrate how the average MAPE and NLPD values evolve over the context length range for all six models. As expected, the MAPE values generally improve with increasing context length as the key features, like the knee-point, begin to become apparent. The pre-trained BNN is a slight exception to this as its values remain very low across the entire context length range for both the simulated and experimental datasets. This strongly suggests that the BNN is able to learn the key features of both datasets and make very accurate predictions, even with only 50\% of the curve visible. The pre-trained sGP performs quite well from 60\% to 80\% context but is prone to significantly overestimating the knee-point at 50\%. The BNN and sGP without pre-training both show significant improvement from 50\% to 80\% context on both datasets and end up converging at similar MAPE values at 80\% context. The large errors for these two models without pre-training at 50\% and 60\% context are likely caused by their larger numbers of parameters which results in overfitting and poorly performing forecasts. 

\begin{figure}[ht]
    \centering
    \includegraphics[width=1\linewidth]{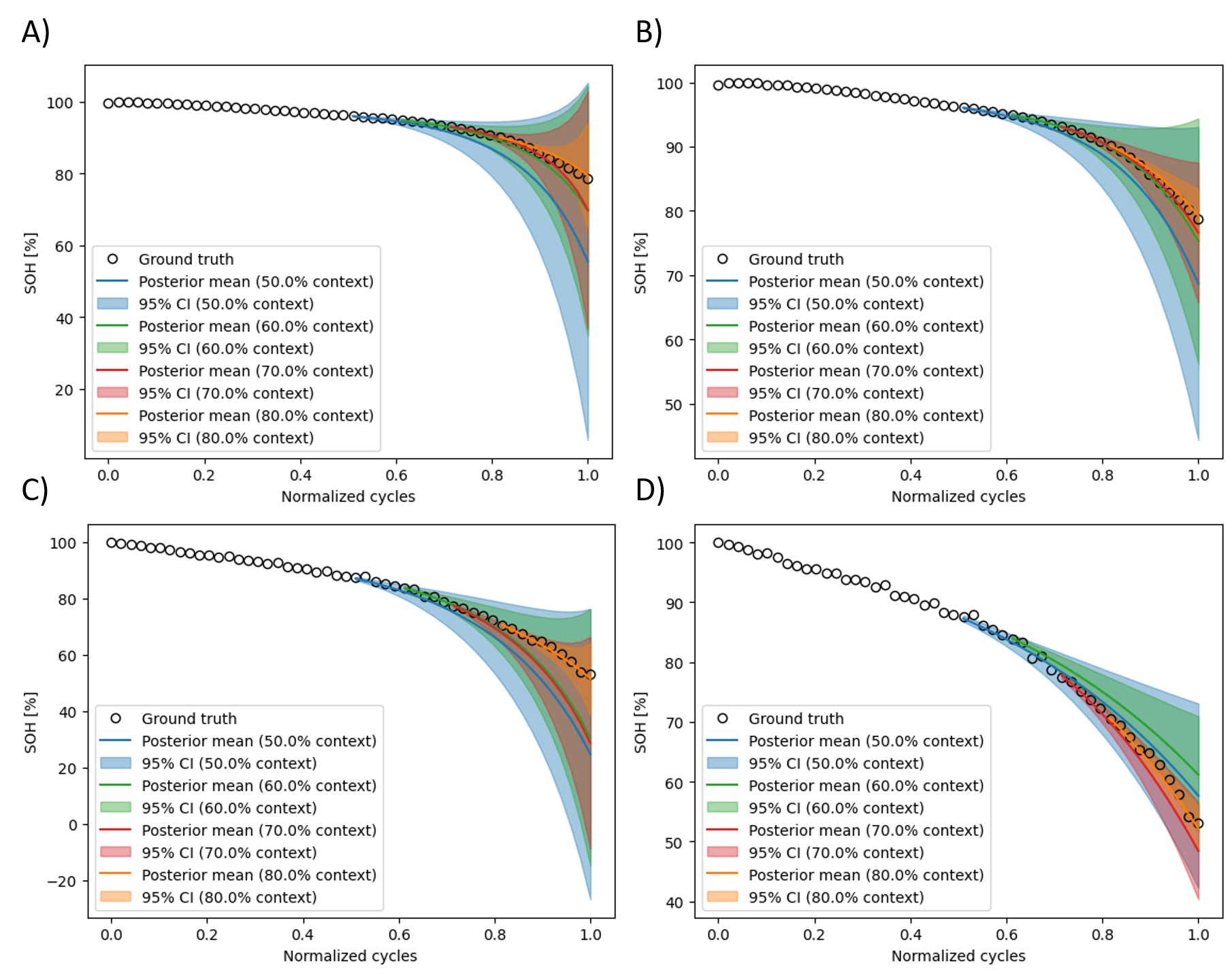}
    \caption{Structured GP posterior mean and 95\% confidence intervals for 50\%-80\% context length without pre-training (A), with pre-training (B) on the experimental dataset and without pre-training (C) and with pre-training (D) on the simulated dataset.}
    \label{fig:context-length-sgp-timeseries}
\end{figure}

Figure \ref{fig:context-length-sgp-timeseries} provides a visual case study for how the sGP model and uncertainty evolve over the context length range without pre-training (A, C) and with pre-training (B, D). In the experimental data sample (A, B) and the simulated data sample (C, D) the models without pre-training steadily improve in accuracy as the context increases. In contrast, the pre-trained samples show highly accurate forecasts across the range with confidence intervals that are mostly consistent with battery physics. 

Overall, it is clear that carrying out long horizon predictions without pre-training can result in significant error and poorly calibrated uncertainty bounds. 

\begin{table}[h]
    \fontsize{8}{12}\selectfont
    \centering
    \begin{tabular}{ |c|c|c|c|c| } 
         \hline
         \rowcolor{lightgray}
         \textbf{Model}& \textbf{50\% context} & \textbf{60\% context} & \textbf{70\% context} & \textbf{80\% context} \\
         \hline
         GP & 5.45\% & 6.08\% & 5.86\% & 3.67\% \\
         \hline
         pt-GP & 6.00\% & 6.35\% & 5.89\% & 3.72\%  \\
         \hline
         sGP & 19.46\% & 6.83\% & 1.59\% & 1.80\% \\
         \hline
         pt-sGP & 7.96\% & 3.13\% & 2.44\% & 1.85\% \\
         \hline
         BNN & 15.81\% & 9.29\% & 5.08\% & 2.42\% \\
         \hline
         pt-BNN & 2.13\% & 1.71\% & 1.47\% & 0.93\% \\
         \hline
    \end{tabular}
    \caption{Average MAPE for all models with and without pre-training on experimental data.}
    \label{tab:MAPE_table_exp}
\end{table}

\begin{table}[h]
    \fontsize{8}{12}\selectfont
    \centering
    \begin{tabular}{ |c|c|c|c|c| } 
         \hline
         \rowcolor{lightgray}
         \textbf{Model}& \textbf{50\% context} & \textbf{60\% context} & \textbf{70\% context} & \textbf{80\% context} \\
         \hline
         GP & -1.25 & -1.22 & -0.02 & 1.12 \\
         \hline
         pt-GP & -1.27 & -1.11 & 0.14 &  1.49 \\
         \hline
         sGP & -0.02 & -0.56 & -1.17 & -1.86 \\
         \hline
         pt-sGP & -0.75 & -1.18 & -1.78 & -2.27 \\
         \hline
         BNN & -0.09  & 4.91 & 11.16 & 10.91 \\
         \hline
         pt-BNN & 0.29 & -0.93 & 0.42 & -2.29 \\
         \hline
    \end{tabular}
    \caption{Average NLPD for all models with and without pre-training on experimental data.}
    \label{tab:NLPD_table_exp}
\end{table}

\section{Conclusions}
We have described how Gaussian processes, structured Gaussian processes and Bayesian neural networks behave with and without pre-training and across a range of context lengths for predicting battery capacity fade. Pre-training has a significant positive impact on both the BNN and sGP models while it shows a minimal improvement for the GP model. The pre-trained BNN achieved the best accuracy but the uncertainty bounds were too narrow, indicating an overly confident model, and the pre-trained sGP does well on both metrics, offering a well-balanced option. 

The approaches outlined in this study provide a flexible framework for accurately predicting battery capacity trajectories with uncertainty by utilizing historical data, domain knowledge and phenomenological models. We expect that applying these techniques to larger datasets with more features i.e. charge energy or cell resistance will yield exciting results and allow for longer predictions using shorter context windows. In addition, the sGP approach allows for direct incorporation of learnable physics-based capacity models and can be pre-trained on past data to help provide deeper physical insight into battery performance predictions while quantifying uncertainty in a principled manner. 

\section*{Acknowledgment}
The development and maintenance of the open-source GPax software package is supported by the Laboratory Directed Research and Development Program at Pacific Northwest National Laboratory, a multiprogram national laboratory operated by Battelle for the U.S. Department of Energy. M.J. Kenney and K.G. Malollari are supported by the energy technology and reliability teams at Amazon Lab 126.

\printbibliography

\end{document}